# A REVIEW OF SOFT ROBOTS

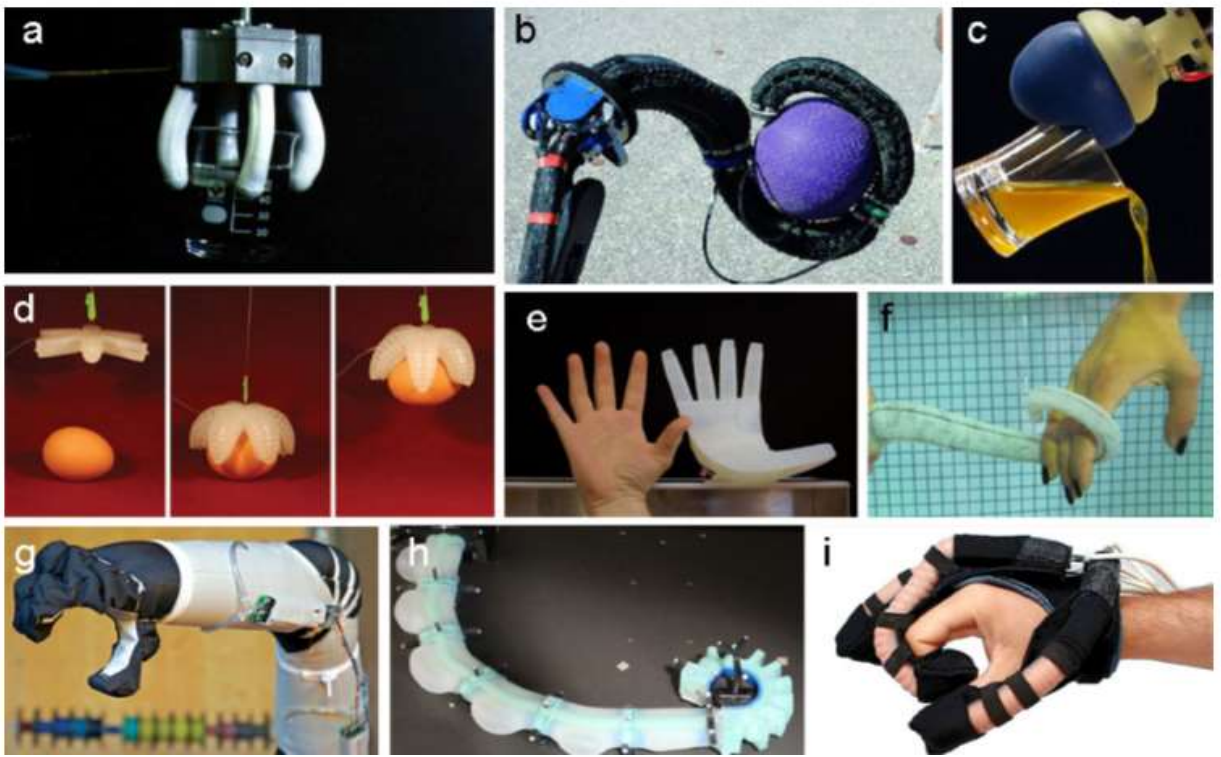

Name: Oladipupo Gideon Gbenga
Email: ggo1@hw.ac.uk;hgrv86@gmail.com

Table of Contents

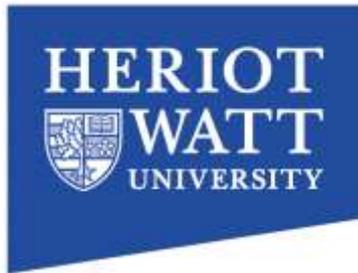

Abstract

Traditional robots have rigid links and structures that limit their ability to interact with the dynamics of their immediate environment. For example, conventional robot manipulators with rigid links can only manipulate objects using specific end effectors. These robots often encounter difficulties operating in unstructured and highly congested environments. A variety of biological organisms exhibit complex movement with soft structures devoid of rigid components. Inspired by biology, researchers have been able to design and build soft robots. With a soft structure and redundant degrees of freedom, these robots can be used for delicate tasks in unstructured environments. This review discusses the motivation for soft robots, their design processes as well as their applications and limitations. Soft robots have the ability to operate in unstructured environment due to their inherent potential to exploit morphological computation to adapt to, and interact with, the world in a way that is difficult with rigid systems. Soft robots could be used for operations, ranging from search and rescue operations in a natural disaster relief effort, and of emerging interest is in the field of medical care as seen in personal robots.

# Introduction

Traditional robots with rigid bodies and links are widely employed in manufacturing and could be coded to perform a specific task effectively and in most cases with certain adaptable constraints [1]. Essentially, traditional robots are built from non-complaint materials [2] and they are constructed mainly from hard materials such as aluminium and steel [3]. Due to inflexibility associated with the rigid links and joints of the robot, they are often unsafe to interact with human beings [1]. In a bid to reduce the related risks while working with traditional robots, the human and robotic workspaces in an industrial environment are often separated [ibid]. The recent Soft robots provide an opportunity to bridge the gap between machines and people [4].

Generally, soft robots are biologically-inspired with inspiration garnered from different capabilities of living organisms such as climbing of geckos, wing flapping motion of butterflies and crawling motion of earthworms amongst others [6]. Soft robots are made from fluids, gels, soft polymers, and other easily deformable matter with little or no rigid material [5]. These materials exhibit similar elastic and rheological properties of soft biological matter and these allow the robot to be more versatile and robust to operate in a constrained environment and more so, for seamless human interaction [ibid]. However, [7] gave a different view on the definition of soft robot and claimed a "soft robot" is appropriately named when the stresses, it is subject to cause it to deform prior to damaging the class of objects for which it was designed. Concisely, a traditional robot could be thought as soft robot when interacting with a harder object, such as a diamond [ibid].

According to [4], soft robotics could be viewed from two main perspectives which are distinguishable from each other. Firstly, the softness of a robot was seen intrinsically due to the features of the robot's body ware which would be the focus of this review [8]. Secondly, softness of a robot was viewed from the soft interaction provided when rigid links are used to control the actuator stiffness of the robot [9]. [10] noted that soft robotics was first used to describe robot with rigid links and mechanically complaints joints with variable stiffness. However, this would not be discussed in detail in this review.

In the first approach, robots are made of soft materials and they undergo high deformations during interaction [8]. Soft actuators and materials which could vary in their stiffness are used in this method, and their control is partially embedded in the body morphology [ibid]. This approach exploited the material properties of the robot and its capacity to interact with the environment [ibid]. However, in the second approach, robots are built with traditional rigid links, but the control system varies the resistance that the robot has to exhibit when interacting with the environment, people and object in this case, either through compliance or impedance control schemes [9].

[1] and [4] further reiterated that soft robots use compliant material and are built using non-traditional manufacturing processes. Moreover, these robots integrate several parts such as actuators and sensors as well as computation amongst others [ibid]. As regards to the assembly process, joining between the materials are made by casting, laminating or adhesives as against the using of bolts and nuts in traditional robots [2]. More specifically, soft robots are viewed as biomimetic robots in that they mimic specific capabilities of certain organisms. However, it is worth noting that soft robots are different from previous biomimetic robots such as snake robots, humanoid robots and robotic fish that are built using traditional manufacturing processes [ibid]. This review of soft robots is organised into fundamentals of soft robots, manufacturing processes of soft robots as well as application of soft robots and its limitations.

## 1   Fundamentals of soft robots

Soft structures made from soft materials and with mobile capabilities abound in nature [8]. For instance, muscular hydrostats such as elephant trunks, mammal and lizard tongues as well as octopus arms as shown in Figure 1 are soft structures that can bend, extend and twist [ibid].

These abilities as applied to soft robots are not just different ways to perform usual robot functions; they are rather capabilities for performing actions such as squeezing, stretching, climbing and growing [10].   These actions would be impossible using a robot design based on rigid links approach.   The integration of soft-matter technologies with morphological computation and embodied paradigms has greatly contributed to an increase in the abilities of soft robots and this is evident in their better performance over traditional robots in real environments [ibid].

Fundamental understanding of the morphology and functionality of soft structures in nature, however, increases insight and can lead to new design concepts in soft robotics [8].   Most of the potential capabilities of soft robots is evident in the natural world [ibid].   Hence, the hydrostatic skeletons and muscular hydrostats would be highlighted.

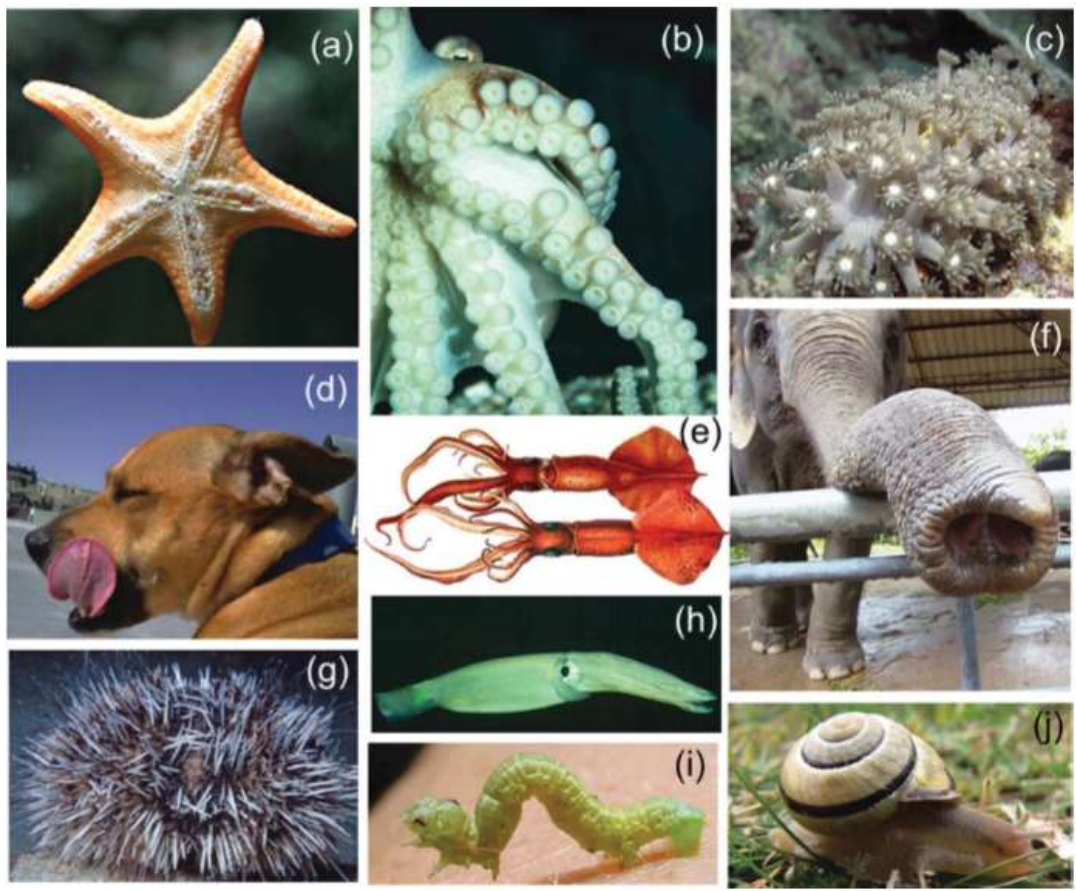

*Figure 1.* Examples of hydro skeletons and muscular hydrostats: (a) tube feet in starfish, (b) octopus arms, (c) colonial anemone, (d) mammalian tongue, (e) squid, (f) elephant trunk, (g) echinoid, (h) Illex illecebrosus, (i) inchworm, and (j) snail feet [8].

## 1.1 Hydrostatic skeletons and muscular hydrostats

Hydrostatic skeletons according to [11] is a fluid mechanism that provides a means by which contractile elements may be antagonized. Hydrostatic skeletons occur in a remarkable variety of organisms with examples not only from invertebrates but also from vertebrates [ibid]. A hydrostatic skeleton is typically considered to include liquid-filled cavity surrounded by a muscular wall reinforced with connective tissue fibres. Comparison of a variety of hydrostatic skeletal support systems reveals that the extent and volume of the liquid-filled cavity is variable [ibid].

Later work particularly identified several hydrostatic skeletons, otherwise known as muscular hydrostats. They consist of a tightly packed three-dimensional array of muscle fibres [ibid]. Examples of muscular hydrostats include the arms and tentacles, fins, suckers and mantles of cephalopod molluscs, a variety of molluscan structures, the tongues of many mammals and lizards, and the trunk of the elephant [ibid].

The basic arrangement of muscle fibres in vermiform hydrostatic skeletons typically includes both circular and longitudinal muscle fibres [ibid]. The hydrostatic skeleton provides a means by which these two muscle fibre orientations can antagonise one another, producing a variety of movements including elongation, shortening, and bending [ibid]. The function of the hydrostatic skeletal support system relies on the fact that the enclosed liquid-filled cavity, typically, coelom or water, is constant in volume [ibid]. In this regard, any decrease in one elongation must result in an increase in another. Consequently, to create the elongation of the body, the circular muscles contract, decreasing the diameter and thereby increasing the length.

In addition to the basic circular and longitudinal muscle arrangements, other muscle fibre arrangements are observed and allow additional types of movement. The four possible categories of movement are elongation, shortening, bending and torsion [ibid]. Elongation requires muscle fibres that are arranged such that their contraction decreases the diameter of the body or organ [ibid]. Shortening occurs because of contraction of the longitudinal musculature and provides a means by which the muscle responsible for elongation may be antagonised [ibid].

Bending movements require contraction of longitudinal muscle along one side of the body or organ. The bending movement is greatest if the longitudinal muscle is peripherally arranged, that is, located as from the central axis as possible [ibid]. Since any decrease in length due to longitudinal compressional force must result in an increase in diameter of a hydrostatic system, muscles arranged to control diameter can provide the resistance to longitudinal compressional force required for bending. It is worth noting that these are the same muscle arrangements that produce elongation. However, in the case of bending, these muscles operate in synergy with the longitudinal muscles rather than antagonistically [ibid].

The last category of movement in hydrostatic skeletons according to [11] is torsion or twisting around the long axis. The muscles responsible for this movement are arranged helically around the body or organ [ibid]. For the torsion in either direction to be possible, both right- and left-handed helical muscle layers must be present. The torsional moment of these muscle layers is maximised if the layers are located toward the outer surface, as far from the central axis as possible [ibid].

These relative movements of musculature depending on the various arrangements result to transmission forces which forms the basis in the design of soft robots and not by the rigid links as in the case of traditional robots [8].

## 2    Manufacturing processes of soft robots

In the field of soft robotics, the lack of reliable and robust soft actuators has been one of the setbacks in developing soft robots [4]. However, the emergence of new technologies in this field is stimulating more research interests. One of these is the development of flexible or soft material which is crucial and fundamental for the development of soft robots [ibid]. A soft robot integrates the sub-systems of a conventional robot which are, actuation system, sensing system, stretchable electronics, a computation and control system, power source as well as modelling and kinematics [1]. The manufacturing steps of the various sub-systems would be highlighted subsequently.

### 2.1    Actuation System

The segments of a soft robot are usually actuated in one of two ways as shown in Figure 1. Firstly, it could be achieved through variable length tendons, in the form of tension cables [1] or Shape Memory Alloy (SMA) actuators. A typical SMA is robotic octopus' arms in which the actuators is embedded in soft segments as shown in Figure 3 [ibid]. Secondly, pneumatic actuation is used to inflate channels in a soft material and cause a desired deformation. Pneumatic Artificial Muscles (PAMs), otherwise called McKibben actuators, are examples of compliant linear soft actuators composed of elastomer tubes in fibre sleeves [ibid]. Moreover, Fluidic Elastomer Actuators (FEAs) followed the aforementioned types and they are highly extensible and adaptable, low-power soft actuator. FEAs comprise synthetic elastomer films operated by the expansion of embedded channels under pressure [ibid].

Furthermore, it is worth noting that most soft robot prototypes have used pneumatic or hydraulic actuation. However, more research efforts in recent times has focused on the development of electrically activated soft actuators composed of electroactive polymers (EAPs). Since energy is typically most readily stored in electrical form, and computation is usually done on electronic circuits, it may be more efficient to directly use electrical potential to actuate soft robots

## 2.2 Sensing System

Sensing in soft robots based on their morphology and compliance require that the sensors must be bendable. This underscore the reason why many traditional sensors such as encoders, metal or semiconductor strain gages, or inertial measurement units (IMUs) amongst others are unsuitable [1]. While flexible-bending sensors based on piezoelectric polymers are available as commercial products, these may not be appropriate due to the need for all elements of the system to be both bendable and stretchable. [12] cited by [1] revealed that soft, stretchable electronics might enable new sensing modalities for soft robots.

## 2.3 Stretchable Electronics

Traditional and rigid electronics has been the basis for storing the control algorithms and connect to the systems' actuators, sensors, and power sources in most of the integrated soft robotics system [1]. However, the recent effort in the field of soft and stretchable electronics is opening a new frontier which would ensure its greater integration with soft robots, resulting in completely soft prototypes. Details of the manufacturing processes for stretchable electronics are outlined in [13].

## 2.4 Computation and Control System

The movements of soft robots cannot be confined to planar motions which contrasts with rigid robots, whose movements can be described by six degrees of freedom (DOF), three rotations and three translations about the x, y, and z axes [1]. Soft materials are elastic in nature and could bend, twist, stretch, compress, buckle, and wrinkle amongst others. Put differently, the motion associated with soft robot could be viewed as presenting an infinite number of DOF, thereby making the control of soft robots very challenging. In this regard, a new algorithm is necessary to control soft robots as well as high-level planning. According to [14] cited in [1], an understanding of the working principles and control of soft organisms, such as the octopus has led to a model for the control of soft robots. Similarly, mobile soft robot is modelled after the motion of caterpillar [15].

## 2.5 Power Sources

A suitable power source for soft robots actuation must be stretchable and portable [1]. For pneumatic actuators, existing fluidic power sources are not soft and are usually big and bulky. Most of the current off the-shelf pressure sources are generally limited to compressors or pumps and compressed air cylinders [16]. Drawing analogy from electrical systems, compressors are like generators as they convert electrical energy into mechanical energy, and compressed gas cylinders are similar to capacitors as they store a pressurized fluid in a certain volume to be discharged when required [1]. It is worth stating that electrically powered actuators as well as the electrical controllers for pneumatic systems require soft, flexible, lightweight electrical power sources [17]. As with soft electronics, this is an area of research that is being currently explored [1].

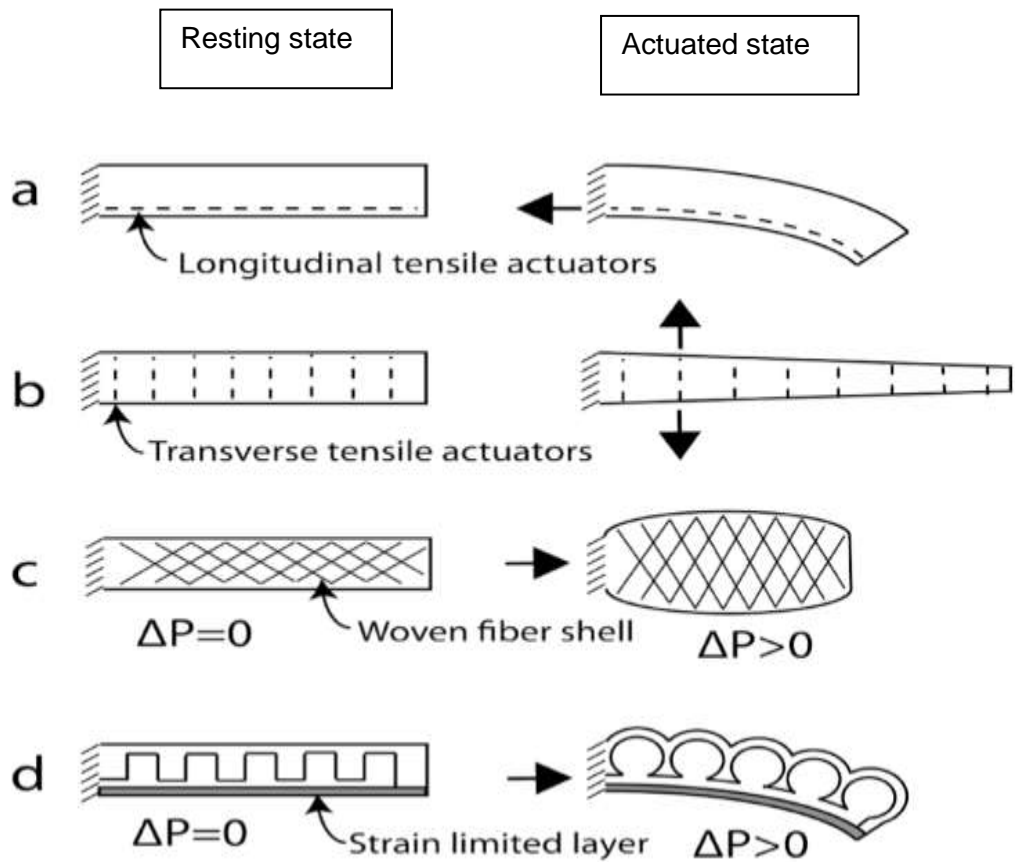

*Figure 2*: Common approaches to actuation of soft robot bodies in resting (left) and actuated (right) states [1]
a. Longitudinal tensile actuators (e.g. tension cables or shape memory alloy actuators which contract when heated) along a soft robot arm cause bending when activated.
b. Transverse tensile actuators cause a soft robot arm to extend when contracted (a muscle arrangement also seen in the octopus72).
c. Pneumatic artificial muscles composed of an elastomeric tube in a woven fibre shell. A pressure applied internally causes the tube and shell to expand radially, causing longitudinal contraction.
d. Fluidic elastic actuator or Pneu-Net design consisting of a pneumatic network of channels in an elastomer that expand when filled with a pressurized fluid, causing the soft body to bend toward a strain limited layer (e.g. a stiffer rubber or elastomer embedded with paper or other tensile fibres

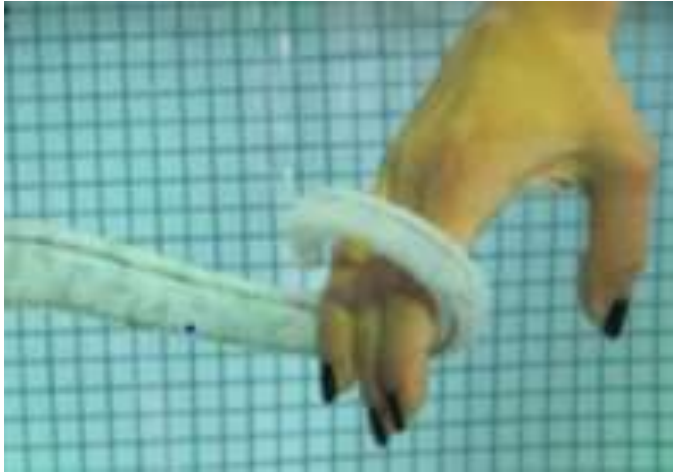

*Figure 3:* Octopus-inspired manipulation [14]

## 2.6  Modelling and Kinematics

As earlier noted, a novel algorithm is required for soft robots as the traditional robots' approach is incompatible with it. In the same vein, the kinematics and dynamics of soft robotic systems differs from those of traditional, rigid-bodied robots [1]. Since its composition is made up of series of actuation elements, these robots approach a continuum behaviour [ibid]. In theory, the final shape of the robot can be described by a continuous function and modelling this behaviour requires continuous mathematics [ibid]. [18] cited in [1] noted that researchers have developed new static, dynamic and kinematic models that capture soft robots' ability to bend and flex.

## 3  Application of soft robots

As previously noted, soft robots have distributed deformation with an infinite number of DOF theoretically. This leads to a hyper-redundant configuration space wherein the robot tip can attain every point in the three-dimensional workspace with an infinite number of robot shapes or configurations [5]. This makes them ideal for applications such as personal robots that interact with people without causing injury, service and painting robots that need high dexterity to reach confined spaces, medical robots, especially for use in surgery, and defence and rescue robots that operate in unstructured environments. A typical application of soft robots in field operations as shown in Figure 4 could range from military reconnaissance field operations, to natural disaster relief search operations as well as oil and gas pipeline monitoring and maintenance [5].

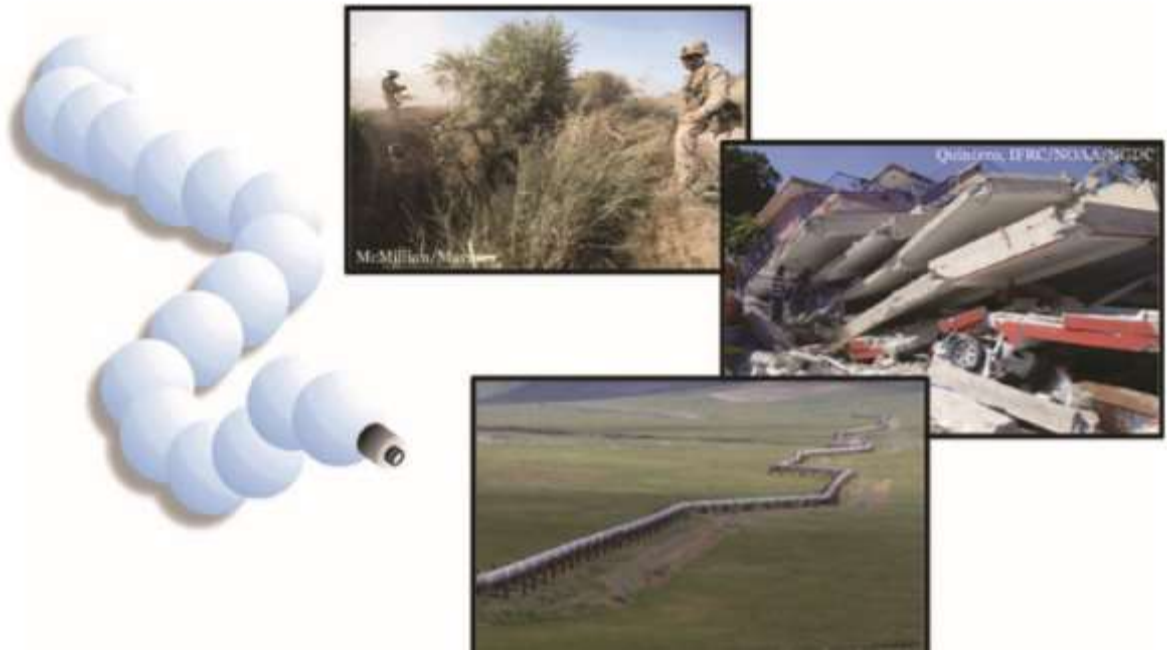

*Figure 4*: Soft field robot for military reconnaissance, natural disaster relief, and pipe inspection [5].

## 4      Limitations of Soft Robots

As earlier stated, conventional machines and robots are not always well suited for human–machine interaction due to their rigidity [5]. Similarly, soft robots are essentially limited by their mechanical compliance and will not be appropriate for applications requiring high power or precision [ibid]. For instance, it is unlikely that soft robots made mainly from fluids and elastomers would replace heavy-duty industrial robots. Likewise, on the small scale, machine precision often requires rigid parts that lock tightly in place and do not slacken or deform elastically when loaded with surface tractions [ibid]. While natural neural tissue is soft and capable of extraordinary computational power, micro-engineered electronics are presently constructed from rigid materials with precisely spaced submicron features [ibid]. Until there is an elastically soft artificial brain, soft robots will require rigid microprocessors for signal processing and actuator control.

## Conclusion

Soft robots inspired by soft body organisms have opened a new frontier in the field of robotics. This has motivated development of new algorithms and techniques in the aspect of its actuation, sensing, computation as well as power sources and kinematics modelling. This is necessary as the approach and algorithm of traditional robots is incompatible with soft robots. Soft robots have the capability to operate in complex and unstructured environment due to their inherent potential to exploit morphological computation to adapt to, and interact with, the world in a way that is difficult or impossible with rigid systems. They could be employed in search and rescue operations in a natural disaster relief effort, and of interest is their ability to interact with people without causing injury especially in the field of medical care as seen in personal robots.

# References

[1] D. Rus and M.T. Tolley. "Design, Fabrication and Control of Soft Robots." Nature 521, no. 7553, pp. 467-475, 2015. (online). Available: http://dx.doi.org/10.1038/nature14543.

[2] K.J. Cho, J.S. Koh, S. Kim, W. S. Chu, Y. Hong and S. H.Ahn . "Review of manufacturing processes for soft biomimetic robots", *Int. J. Precis. Eng. Manuf.* (2009) 10: 171. (online). Available: https://doi.org/10.1007/s12541-009-0064-6.

[3] N. W. Bartlett, M. T. Tolley, J. T. B. Overvelde, J. C. Weaver, B. Mosadegh, K. Bertoldi, G. M. Whitesides, and R. J. Wood. "A 3D-printed, functionally graded soft robot powered by combustion", *SOFT ROBOTICS.* (online). Available: http://science.sciencemag.org.

[4] C. Laschi and M. Cianchetti. "Soft robotics: new perspectives for robot bodyware and control" (online). Available: https://doi.org/10.3389/fbioe.2014.00003.

[5] C. Majidi. "Soft Robotics: A Perspective—Current Trends and Prospects for the Future", *Soft Robotics*, vol.1, no. 1. 2013. (online). Available: http://sml.me.cmu.edu/files/papers/majidi_soro2013.pdf

[6] S. Kim, C. Laschi and B. Trimmer, "Soft robotics: a bioinspired evolution in robotics", *Trends in Biotechnology*, vol. 31, iss. 5, pp 287-294, 2013. (online). Available: https://www.sciencedirect.com/science/article/pii/S0167779913000632.

[7] P. Polygerinos, N. Correll, S. A. Morin, B. Mosadegh, C. D. Onal, K. Petersen, M. Cianchetti, M. T. Tolley and R. F. Shepherd. "Soft Robotics: Review of Fluid-Driven Intrinsically Soft Devices; Manufacturing, Sensing, Control, and Applications in Human-Robot Interaction", *Advanced Engineering Materials*, vol. 19, no. 12, 2017. (online). Available: https://onlinelibrary.wiley.com/doi/epdf/10.1002/adem.201700016

[8] D. Trivedi, C. D. Rahn, W. M. Kier, and I. D. Walker, "Soft Robotics: Biological Inspiration, State of the Art, and Future Research," *Applied Bionics and Biomechanics*, vol. 5, no. 3, pp. 99-117, 2008. (online). Available: https://doi.org/10.1080/11762320802557865.

[9] A. Albu-Schäffer , O. Eiberger, , M. Grebenstein, S. Haddadin, , C. Ott, T.Wimbock, S. Wolf and G. Hirzinger, Soft robotics. *IEEE Robot. Autom. Mag.* 15, pp 20–30, 2008. (online). Available: doi: 10.1109/MRA.2008.927979.

[10] C. Laschi, B. Mazzolai, and M. Cianchetti. "Soft Robotics: Technologies and Systems Pushing the Boundaries of Robot Abilities." *Science Robotics*" vol. 1, iss.1, 2016. (online). Available: https://robotics.sciencemag.org/content/1/1/eaah3690.

[11] W.M. Kier. "Hydrostatic skeletons and muscular hydrostats", *In Biomechanics (Structures and Systems): A Practical Approach*, edited by Biewener, AA. Oxford: IRL Press, pp 205–231, 1992. (online). Available: http://labs.bio.unc.edu/kier/pdf/kier_1992.pdf.

[12] S. Yang and N. Lu, "Gauge factor and stretchability of silicon-on-polymer strain gauges" *Sensors*, vol. 13, pp 8577-8594, 2013. (online). Available: https://www.mdpi.com/1424-8220/13/7/8577.

[13] J. A. Rogers, T. Someya and Y. Huang, "Materials and mechanics for stretchable electronics".*Science*. vol. 327, pp1603-1607, 2010. (online). Available: http://muri.engr.utexas.edu/sites/default/files/publication/10-18%20sciencerev.pdf.

[14] L. Margheri, C. Laschi, and B. Mazzolai, Soft robotic arm inspired by the octopus: I. from biological functions to artificial requirements. *Bioinspiration & biomimetics,* vol. 7, no. 2 p.025004 (12pp), 2012 (online). Available: https://iopscience.iop.org/article/10.1088/1748-3182/7/2/025004/meta

[15] H.T. Lin, G. G. Leisk, and B. Trimmer, "GoQBot: a caterpillar-inspired soft-bodied rolling robot". *Bioinspiration & biomimetics*, vol. 6, no.2*,* p.026007 (14pp), 2011. (online). Available: https://iopscience.iop.org/article/10.1088/1748-3182/6/2/026007/meta.

[16] M. Wehner, T. T. Michael, M. Yiğit, P. Yong-Lae, M. Annan, D. Ye, O. Cagdas, F. S. Robert, M. W. George and J. W. Robert. "Pneumatic energy sources for autonomous and wearable soft robotics" *Soft Robotics* vol. 1, pp. 263–274 2014. (online). Available: https://www.liebertpub.com/doi/10.1089/soro.2014.0018.

[17] X. Sheng, Y. Zhang, J. Cho, J. Lee, X. Huang, L. Jia, J. A. Fan, Y. Su, J. Su, H. Zhang, H. Cheng, B. Lu, C. Yu, C. Chuang, T. Kim, T. Song, K. Shigeta, S. Kang, C. Dagdeviren, I. Petrov, P. V. Braun, Y. Huang, U. Paik, and J. A. Rogers. "Stretchable Batteries with Self-similar Serpentine Interconnects and Integrated Wireless Recharging Systems." *Nature Communications,* vol. 4, p 1543, 2013. (online). Available: https://www.nature.com/articles/ncomms2553.

[18] R. J., Webster and B. A. Jones, "Design and kinematic modelling of constant curvature continuum robots: A review." *The International Journal of Robotics Research,* vol. 29, iss. 13 pp 1661-1683, 2010. (online). Available: https://journals.sagepub.com/doi/abs/10.1177/02783649103681